\documentclass[11pt]{article}
\usepackage{acl2016}
\usepackage{times}
\usepackage{latexsym}
\usepackage{amsmath}
\usepackage{graphicx}
\usepackage[english]{babel}
\usepackage[utf8x]{inputenc}
\usepackage[T1]{fontenc}

\aclfinalcopy % Uncomment this line for the final submission
\title{Topic Based Sentiment Analysis Using Deep Learning }

\author{Sharath T. S.\\
	    University of Santa Cruz\
	    Santa Cruz \\
	    CA, USA\\
	    {\tt sturuvek@ucsc.edu}
	  \And
	Shubhangi Tandon\\
  	University of Santa Cruz\
	Santa Cruz \\
	CA, USA\\
	{\tt shtandon@ucsc.edu}}

\date{}

\begin{document}

\maketitle

\begin{abstract}
 In this paper , we tackle Sentiment Analysis conditioned on a Topic in Twitter data using Deep Learning . We propose a 2-tier approach : In the first phase we create our own Word Embeddings and see that they do perform better than state-of-the-art embeddings when used with standard classifiers. 
 We then perform inference on these embeddings to learn more about a word with respect to all the topics being considered, and also the top n-influencing words for each topic. 
 In the second phase we use these embeddings to predict the sentiment of the tweet with respect to a given topic, and all other topics under discussion. 
\end{abstract}

\section{Introduction}

Twitter is a commonly used microblogging platforms where millions of users express their opinions on various topics or domains. These opinions can be regarding a political issue, technology, sports and entertainment, a particular product, a celebrity or any person of interest. 

Due to the sheer volume of data, it is impossible to manually go through the tweets to extract opinions and sentiments. Hence, there arises a need for automated systems that extract polarity of an opinion with respect to a particular topic. 

It is vital that the context with respect to a topic is understood since words and phrases have have sentiments under different contexts. In it is clearly seen that the word "court" has radically different sentiment with respect to a political and a sporting scenario. All these subtleties and nuances have to be addressed for the task of topic/entity based sentiment prediction. In the rest of the paper, we will use the terms topic and entity interchangeably.  

We solve this problem by creating our own word embedding and then using these embedding for sentiment classification. 
In this paper, we have also tried to demystify the word embedding representing the sentiment it contributes for every topic (see figure \ref{fig:cnn} ).  We have introduced a model , where it is possible to make positive and negative sentiment nuclei for each topic ( figure \ref{fig:amazon} ) . 
\section{Related Work}
Social media , especially micro-blogging websites like twitter are a mine of unstructured but very vocal data. People take to these platforms to express their views and sentiments on several topics.  

There has been a lot of work on topic based sentiment analysis particularly using recurrent neural networks because of its recent success on language\cite{nallapati2016summarunner} , \cite{cheng2016neural}, \cite{bahdanau2014neural}. Twitter based word embeddings have been used in \cite{xiong2016improving} on a few million tweets. There has also been focus on input topic embeddings to network and also attention over sentence level nodes to extract information with respect to the topic \cite{wangattention} \cite{jebbara2016aspect}.

There is also feature engineered topic extraction using information from POS trees to distribute probability mass across every work in a sentence with respect to an topic. Distance embeddings have also been used to give weights to each word in a sentence according to it's position with respect to the topic or topic of interest\cite{jebbara2016aspect}. There has also been focus on identifying aspects in a sentence and predict the sentiment on those aspects \cite{wangattention}\cite{dhanush2016aspect}.

There has also been focus on methods that make use of LDA, BTM topic modeling approaches and then use semantic and syntactic approaches to classify sentiment based on aspect \cite{pavlopoulos2014aspect}. Also most of the approaches are restricted to a small number of domains or a particular domain \cite{gulaty2016aspect} \cite{thet2010aspect} and don't perform well generally on other topics. 

All of these approaches either require hand engineering with respect to an extent topics, POS tags, etc., or completely lack interpretability.

The model we propose is uses the custom embeddings generated to predict the sentiment conditioned on the tweet and the topic. The CNN model is interpretable in the sense that it highlights the sentiment contribution of each word with respect to every topic. Also,  given a topic, we can establish it's  positive and neutral aspects by visualizing words or phrases with the most positive or negative contributions in the predictions obtained from phase 1.  
\begin{figure*}
\centering
\includegraphics[width=\linewidth, height=\textheight,keepaspectratio]{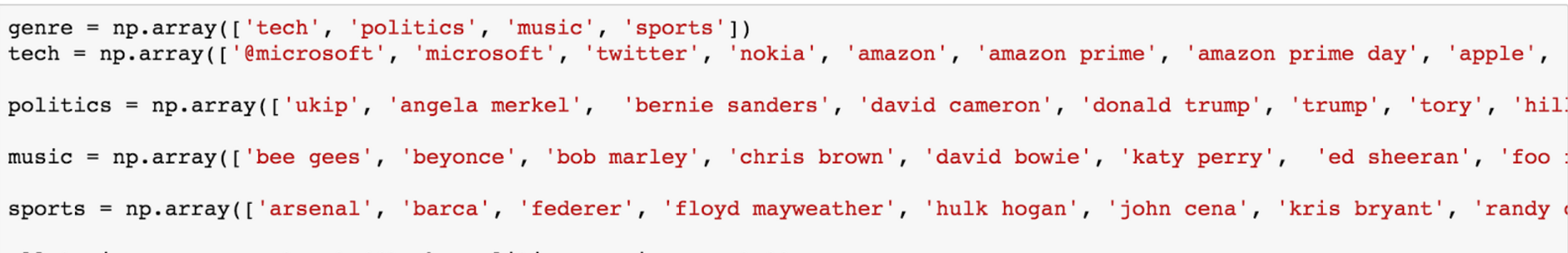}
\caption{\label{fig:data}Sample topics that were selected }
\end{figure*} 

\section{Data}
Our data comes from SemEval 2016 Task 4 (Sentiment Analysis in Twitter) as well as from the Sanders data-set. 
Initially our data has the following attributes :
\begin{enumerate}
\item Tweet id
\item Tweet
\item Topic 
\item Sentiment Score (from -2 to +2)
\end{enumerate}

For the scope of our testing , we picked topics from 4 domains : technology , sports, politics and music. Overall, we selected data from 77 topics (see Figure \ref{fig:data} for sample topics we used) . 

We performed standard preprocessing on the tweets including cleaning htmls, removing mentions , trailing hashtags , removing special characters and punctuations. 
For our custom embedding , we also filter out all words below a minimum occurrence frequency (currently 3) 

We also noticed that our data was heavily skewed towards Positive and Neutral classes, there wasn't enough data for the Negative class. Because of this, the model would have a tendency to over-fit to Positive and Neutral classes. To overcome this , we randomly filtered one-third of data from the Positive and Neutral classes to have a more balanced data distribution . 
At the end of the preprocessing :
\begin{itemize}
\item vocabulary :  14k words
\item genre (domains) :  10
\item number of topics :  370
\item number of tweets :  16895
\end{itemize}

We trained our model on 13300 tweets, cross validated on 700 tweets tested on remaining 2895 tweets. 

\begin{figure*}
\centering
\includegraphics[width=\linewidth, height=\textheight,keepaspectratio]{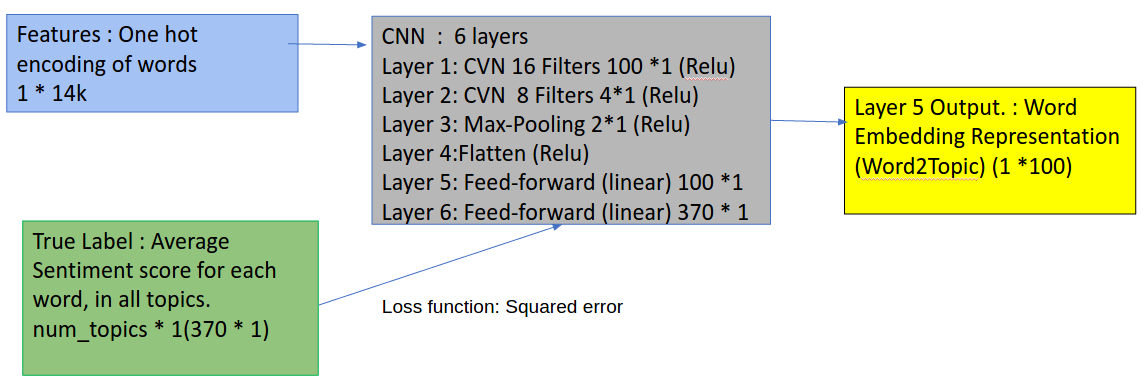}
\caption{\label{fig:word2topic}Diagrammatic representation of word2Topic architecture}
\end{figure*} 

\section{Custom Word Embedding}
\subsection{The Model}
\label{sect:model}
The model to create the custom word embedding is depicted in Figure \ref{fig:word2topic}. 

The first step is obtaining our custom word embeddings that represent the sentiment of a word towards every topic. If the number of topics are \textbf{t} then the we can express each words as a 1 x t vector . 

We made the decision to use custom word to embeddings as opposed to standard embeddings as we found that that the latter over-fit (Subsection~\ref{ssec:baselines}). 
We obtain the custom word embeddings (henceforth called word2topic) by training a CNN . The input to the CNN is a stacked array where each row is the one-hot encoding for every word in our vocabulary .  If our vocabulary had {\textbf{n}} words, then the input would be an array for size n x n.  

The labels to the CNN are a normalized mean of the sentiment expressed by a word with respect to every topic. For every word, the label vector is of size 1 x t. The label array is of size n x t.  

The inputs and labels are fed into a 6 layer CNN (2D). We take inspiration from the architecture of word2vec \cite{mikolov2013distributed} and use the representation from the penultimate hidden layer (FeedForward) for computation of Sentiment with respect to Topic (see section {\ref{sec:pred}) . We use Adam to for optimization and \textit{mean squared error} loss function.

The output from the last layer ( of the same size as label , 1 x t for every word ) is used to perform inference ( subsection \ref{ssec:infer_word_topic}).

\begin{figure*}
\centering
\includegraphics[width=\linewidth, height=\textheight,keepaspectratio]{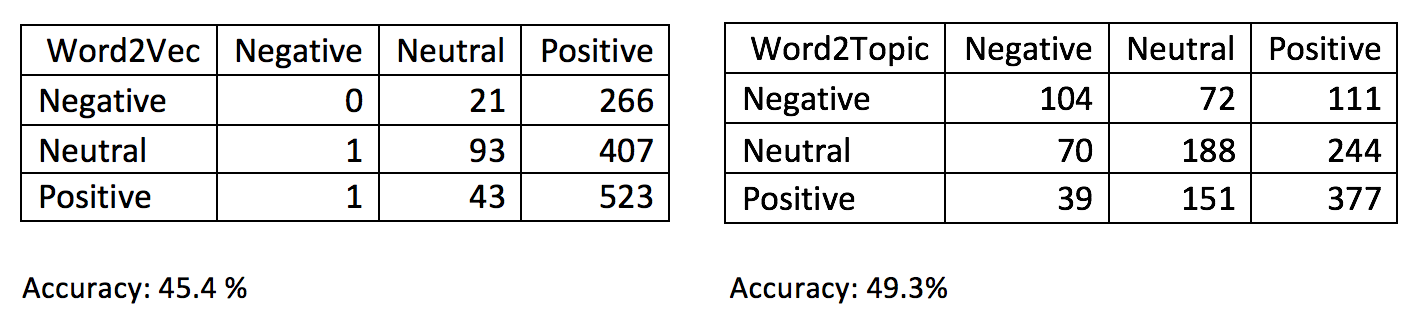}
\caption{\label{fig:base}Confusion Matrix for Logistic Regression of Test Data for Word2Vec and Word2Topic embedding
}
\end{figure*} 

\subsection{Baselines}
\label{ssec:baselines}
The task that we're performing can be split into two parts : the first is to generate custom word embedding for each word that reflects the sentiment of a word with respect to each topic. The second is to predict the sentiment of a tweet with respect to all topics, not just the topic we were given labeled data for. 

Since the first task is pretty standard, we compared our word embedding against state-of-the-art word2vec embedding to train a Logistic Regression Classifier. The results can be seen in figure \ref{fig:base}. Each tweet is represented as the concatenated word embedding representation of each word (padded up to 30 words per sentence) and one-hot encoding of the topic. The label is the sentiment score . 

We see that not only does word2topic give a better accuracy than word2vec, but also that word2vec tends to over-fit the data as it classifies most of the tweets as neutral/positive and hardly classifying anything in the negative class. Word2Topic , performs better and its confusion matrix is stronger on the diagonal. This gives us the motivation to move forward with word2topic as our word embedding for classification .

\begin{figure*}
\centering
\includegraphics[width=\linewidth, height=\textheight,keepaspectratio]{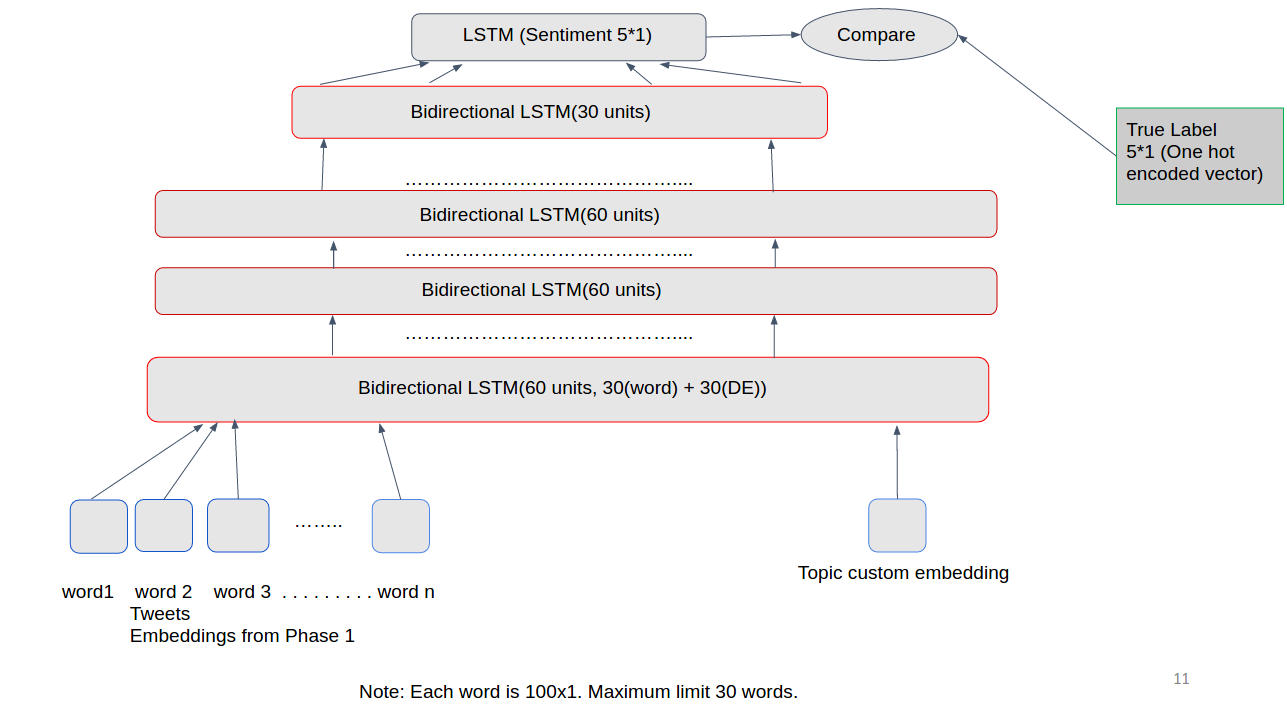}
\caption{\label{fig:rnn}Diagrammatic representation of Sentiment Classification RNN architecture}
\end{figure*} 

\begin{figure*}
\centering
\includegraphics[width=\linewidth, height=\textheight,keepaspectratio]{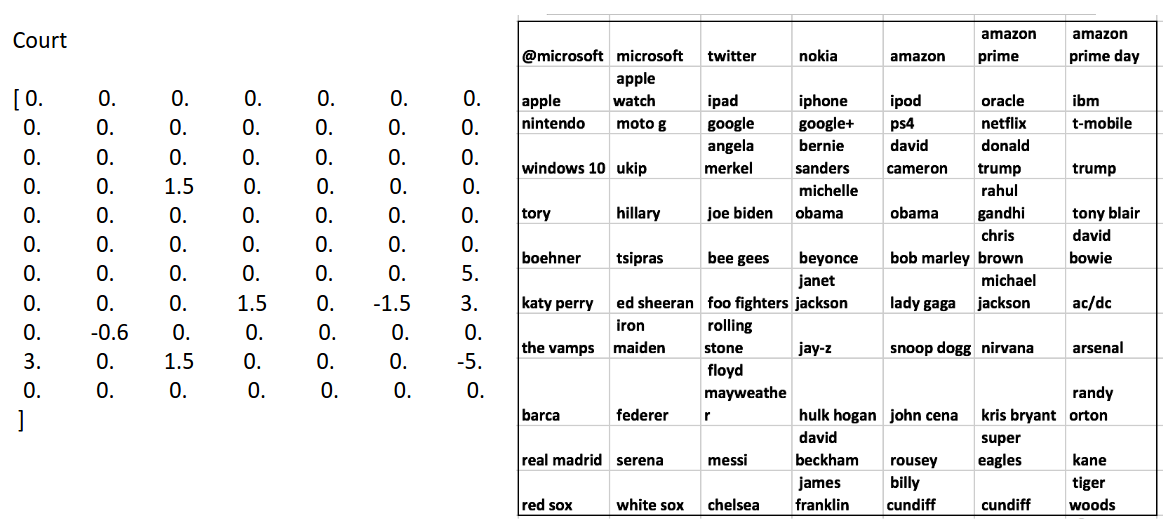}
\caption{\label{fig:cnn} Topic-wise Representation of the word 'court' for Inference }
\end{figure*} 

\begin{figure*}
\centering
\includegraphics[width=\linewidth, height=8 cm,keepaspectratio]{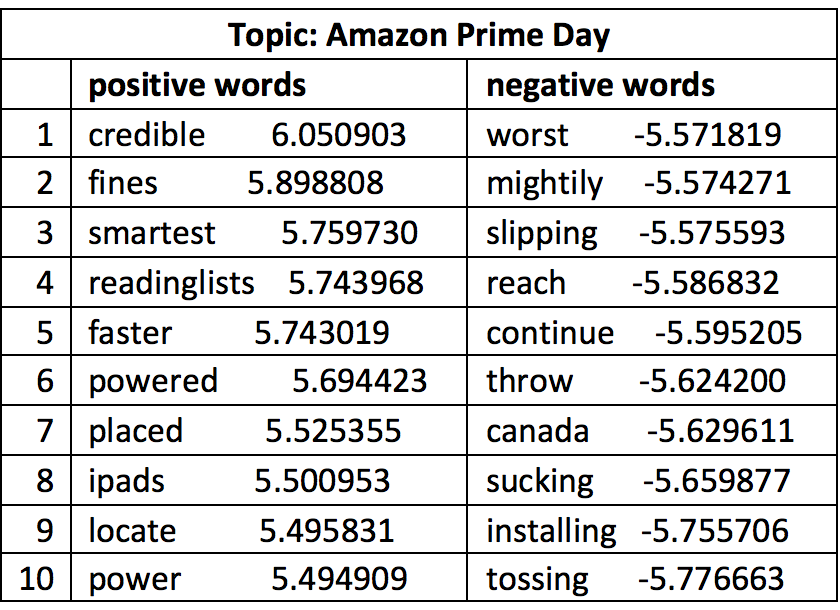}
\caption{\label{fig:amazon} Top influencing words for the topic 'Amazon Prime Day' }
\end{figure*} 

\section{Sentiment Classification}
\label{sec:pred}
We classify topic sentiment at the sentence level and treat it as a conditional modeling problem of predicting the sentiment score, whether positive or negative or neutral, given a tweet text 't' and an topic/topic 'a'. Both these variables serve as input to a model which further classifies the sentiment as either positive, negative or neutral.

Each word in a tweet is represented as the value of the hidden layer of size 100 extracted from phase 1. The topic is represented as a single word embedding. We use a Bidirectional Recurrent Neural network architecture with a LSTM cell for the proposed task and the architecture of the network is shown in \ref{fig:rnn}.  The network can be decomposed into two blocks, the sentence block and the topic block. The sentence block consists of 30 LSTM units which takes in a sentence as input where each sentence is represented as a vector of word embeddings.  The topic block takes in as input the word embedding of the topic and the output of both these blocks are concatenate together in order for the sentence level information and the topic information to interact. We include two additional Bidirectional LSTM layers to increase the model complexity and have a 3 node softmax activation as the output layer for classification. Adam was used as the optimizer with a learning rate of 0.0005. It was trained with a batch size of 64 for 40 epochs with {categorical cross entropy} as the loss function.

\subsection{Results}
\label{ssec:result}
The results from the RNN are shown in \ref{fig:rnn_results}: 
The RNN gives an accuracy of 74.4\% accuracy on 3 class classification \ref{fig:test3} and 64.81\% on 5 class classification\ref{fig:test5}. It can be clearly seen from the confusion matrix that the classifies accuracy identifies negative opinions from the positive ones with a very high accuracy. It misses the fine line between neutral-positive and neutral-negative on a few cases. The training accuracy is also shown in \ref{fig:train5} and \ref{fig:train3} .Hence, future work can be focused on classifiers that first classify neutral comments from opinionated ones, and as a next step, classify positive opinions from the negatives.

\begin{figure*}
\centering
\includegraphics[width=\linewidth, height=5cm,keepaspectratio]{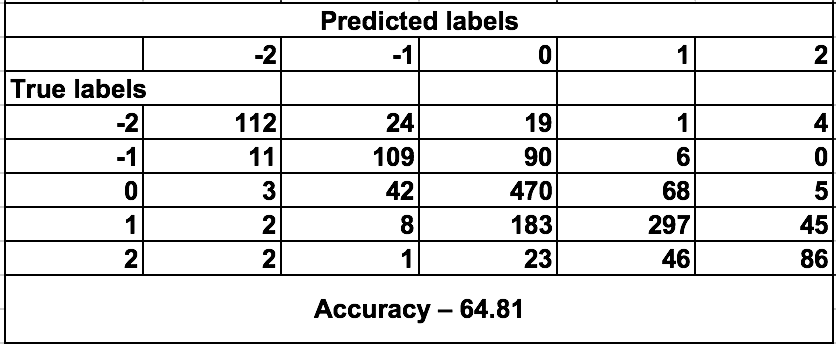}
\caption{\label{fig:test5}Confusion Matrix(Test) for RNN 5 class
}
\end{figure*} 

\begin{figure*}
\centering
\includegraphics[width=\linewidth, height=4cm,keepaspectratio]{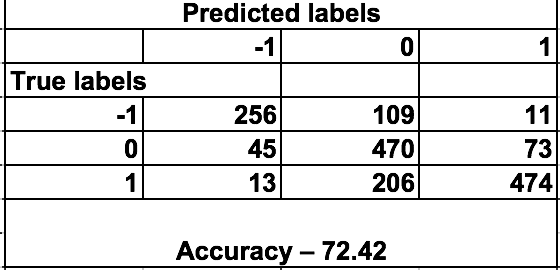}
\caption{\label{fig:test3}Confusion Matrix(Test) for RNN 3 class
}
\end{figure*}

\begin{figure*}
\centering
\includegraphics[width=\linewidth, height=5cm,keepaspectratio]{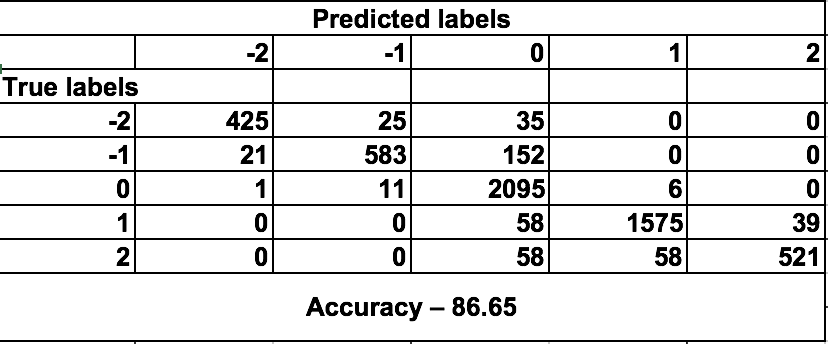}
\caption{\label{fig:train5}Confusion Matrix(Train) for RNN 5 class
}
\end{figure*} 

\begin{figure*}
\centering
\includegraphics[width=\linewidth, height=4cm,keepaspectratio]{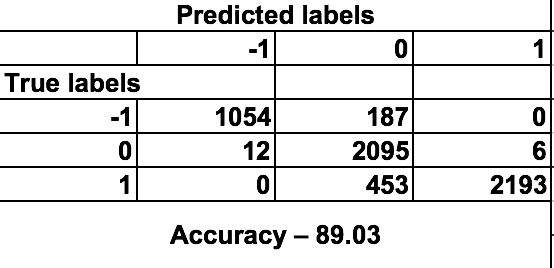}
\caption{\label{fig:train3}Confusion Matrix(Train) for RNN 3 class
}
\end{figure*}  

\section{Inference}
\label{sec:infer}
One of the novel features of our model is that is interpretable. By simply examining the output vectors at each phase , we can perform different kind of inferences . 

\subsection{Sentiment contributed by a word given topic}
\label{ssec:infer_word_topic}
If we look at the output of the CNN model as discussed in section \ref{sect:model}, we see that for each word , we get a 1 x t vector , where t is the number of topics under discussion . For example , we can see the word-embedding of the word 'court' for 77 topics in figure \ref{fig:cnn} .  The arrays have been reshaped to 11 x 7 for better visibility . 

If one were to examine this matrix carefully , we can see that the word court has a positive sentiment for the topic 'angela markel' (1.5) and also for 'real madrid' .  Although the context in which the word court is used for these topics is completely different, yet we are able to capture the sentiment that the word is contributing to with respect to each topic. 

Another way to interpret this would be to think of these values as potential functions , and that the sentiment of a tweet is proportional to the (weighted) sum of the potential functions of all its constituent words. Therefore , the presence of the word 'court' in a tweet about the topic 'real madrid' is going to have a positive effect on the sentiment .

This representation helps us interpret the sentiment contributed by every word in the vocabulary for every topic. 

\subsection{Most influential words for a topic}
\label{ssec:infer_topic_infl_word}
By doing some analysis on our data, we can also infer given any topic, the words that bear the most positive and negative sentiment for it. For example, let’s consider the topic 'Amazon Prime Day' as shown in figure \ref{fig:amazon}. 

Currently we’ve obtained the most positive and negative words for a topic by examining the sentiment expressed by them in the custom word embedding. Words like 'credible' ,'smartest', 'faster' contribute to a very positive sentiment , whereas words like 'worst', 'throw', 'slipping' contribute to a very negative sentiment . The number associated with the word can be interpreted as the potential function of a word given a topic. 

By performing further analysis on these 'influential key words' one can create positive and negative nuclei for every topic. Although, we haven’t yet used these nuclei to refine our results, we would want to incorporate them in future work.

\section{Conclusion}
To conclude, we have built a model for sentiment classification of tweets given topics using deep learning. We do this in two parts : first by creating custom word embedding , second by predicting using custom embedding .  Our model is interpretable , as we can infer the sentiment that a word contributes to every topic and the most influential words for any topic. 
Our model is novel  as given any tweet , it can predict the sentiment for not just one but many topics.

Although, we have created a model that gives us interpretable results, our macro F-score can still be improved. We plan to do this by introducing POS tagging weight on labels when generating word embeddings. We also plan to introduce a sophisticated distance embedding for each tweet from a given topic (again , incorporating POS tagging) .

\section*{Acknowledgments}

We would like to acknowledge the efforts of our course instructor Amita and fellow classmates , who gave us critical advise and guided us to complete our project on time.

\bibliography{acl2016}
\bibliographystyle{acl2016}

\end{document}